\documentclass[sigconf]{acmart}

\usepackage{svg}
\usepackage{mfirstuc}
\usepackage[utf8]{inputenc}
\usepackage{balance} 

\usepackage{afterpage}

\usepackage{makecell}
\usepackage{subcaption}
\usepackage{xcolor}
\usepackage{colortbl}
\usepackage{multirow}
\usepackage{graphicx}
\usepackage{subcaption}
\usepackage{amsmath}
\usepackage{booktabs}
\usepackage{listings}
\usepackage{array}
\usepackage{amssymb}  
\usepackage{arydshln}
\usepackage{multirow}

\usepackage{graphicx}
\usepackage{hyperref}

\AtBeginDocument{%
  }

\copyrightyear{2025}
\acmYear{2025}
\setcopyright{acmlicensed}
\acmConference[MM '25] {Proceedings of the 33rd ACM International Conference on Multimedia}{October 27--31, 2025}{Dublin, Ireland.}
\acmBooktitle{Proceedings of the 33rd ACM International Conference on Multimedia (MM '25), October 27--31, 2025, Dublin, Ireland}
\acmISBN{10.1145/3746027.3758248}
\acmDOI{979-8-4007-2035-2/2025/10}

\begin{document}

\title{Beyond the Individual: Introducing Group Intention Forecasting with SHOT Dataset}

\settopmatter{authorsperrow=4}
\author{Ruixu Zhang}
\authornote{Co-first authors.}
\orcid{0009-0004-9985-0204}

\affiliation{%
  \institution{National Engineering Research Center for Multimedia Software, Institute of Artificial Intelligence, School of Computer Science, \\Wuhan University}
  \city{Wuhan}
  \country{China}
  }
  \affiliation{%
  \institution{Tsinghua University}
  \city{Shenzhen}
  \country{China}
  }

\author{Yuran Wang}
\orcid{0000-0002-3065-2830}
\authornotemark[1]
\affiliation{%
  \institution{School of Mathematical Sciences, Peking University}
  \city{Beijing}
  \country{China}
 }
\affiliation{%
  \institution{National Engineering Research Center for Multimedia Software, School of Computer Science, \\Wuhan University}
  \city{Wuhan}
  \country{China}}

\author{Xinyi Hu}
\orcid{0009-0002-2252-8589}
\authornotemark[1]
\affiliation{%
  \institution{National Engineering Research Center for Multimedia Software, Institute of Artificial Intelligence, School of Computer Science, \\Wuhan University}
  \city{Wuhan}
  \country{China}}

\author{Chaoyu Mai}
\orcid{0009-0002-2252-8589}
\affiliation{%
   \institution{National Engineering Research Center for Multimedia Software, Institute of Artificial Intelligence, School of Computer Science, \\Wuhan University}
  \city{Wuhan}
  \country{China}}

\author{Wenxuan Liu}
\orcid{0000-0002-4417-6628}
\affiliation{%
  \institution{School of Computer Science, Peking University}
  \institution{State Key Laboratory for Multimedia Information Processing, \\Peking University}
  \city{Beijing}
  \country{China}}

\author{Danni Xu}
\orcid{0000-0001-9482-0111}
\affiliation{%
  \institution{School of Computing, National University of Singapore}
  \country{Singapore}
}

\author{Xian Zhong}
\orcid{0000-0002-5242-0467}
\affiliation{%
  \institution{Hubei Key Laboratory of Transportation Internet of Things, School of Computer Science and Artificial Intelligence, Wuhan University of Technology}
  \city{Wuhan}
  \country{China}
}

\author{Zheng Wang}
\orcid{0000-0003-3846-9157}
\authornote{Corresponding author. E-mail: wangzwhu@whu.edu.cn.}
\affiliation{%
  \institution{National Engineering Research Center for Multimedia Software, Institute of Artificial Intelligence, School of Computer Science, \\Wuhan University}
  \city{Wuhan}
  \country{China}}

\renewcommand{\shortauthors}{Ruixu Zhang et al.}

\begin{abstract}
Intention recognition has traditionally focused on individual intentions, overlooking the complexities of collective intentions in group settings.
To address this limitation, we introduce the concept of group intention, which represents shared goals emerging through the actions of multiple individuals, and Group Intention Forecasting (GIF), a novel task that forecasts when group intentions will occur by analyzing individual actions and interactions before the collective goal becomes apparent.
To investigate GIF in a specific scenario, we propose SHOT, the first large-scale dataset for GIF, consisting of 1,979 basketball video clips captured from 5 camera views and annotated with 6 types of individual attributes.
SHOT is designed with 3 key characteristics: \textbf{multi-individual information}, \textbf{multi-view adaptability}, and \textbf{multi-level intention}, making it well-suited for studying emerging group intentions.
Furthermore, we introduce GIFT (Group Intention ForecasTer), a framework that extracts fine-grained individual features and models evolving group dynamics to forecast intention emergence.
Experimental results confirm the effectiveness of SHOT and GIFT, establishing a strong foundation for future research in group intention forecasting. The dataset is available at https://xinyi-hu.github.io/SHOT\_DATASET.

\end{abstract}

\vspace{-8mm}

\ccsdesc[500]{Computing methodologies~Activity recognition and understanding}

\keywords{Group Intention; Group Intention Forecasting; Intention Recognition}

\begin{teaserfigure}
    \centering
  \includegraphics[width=1\textwidth]{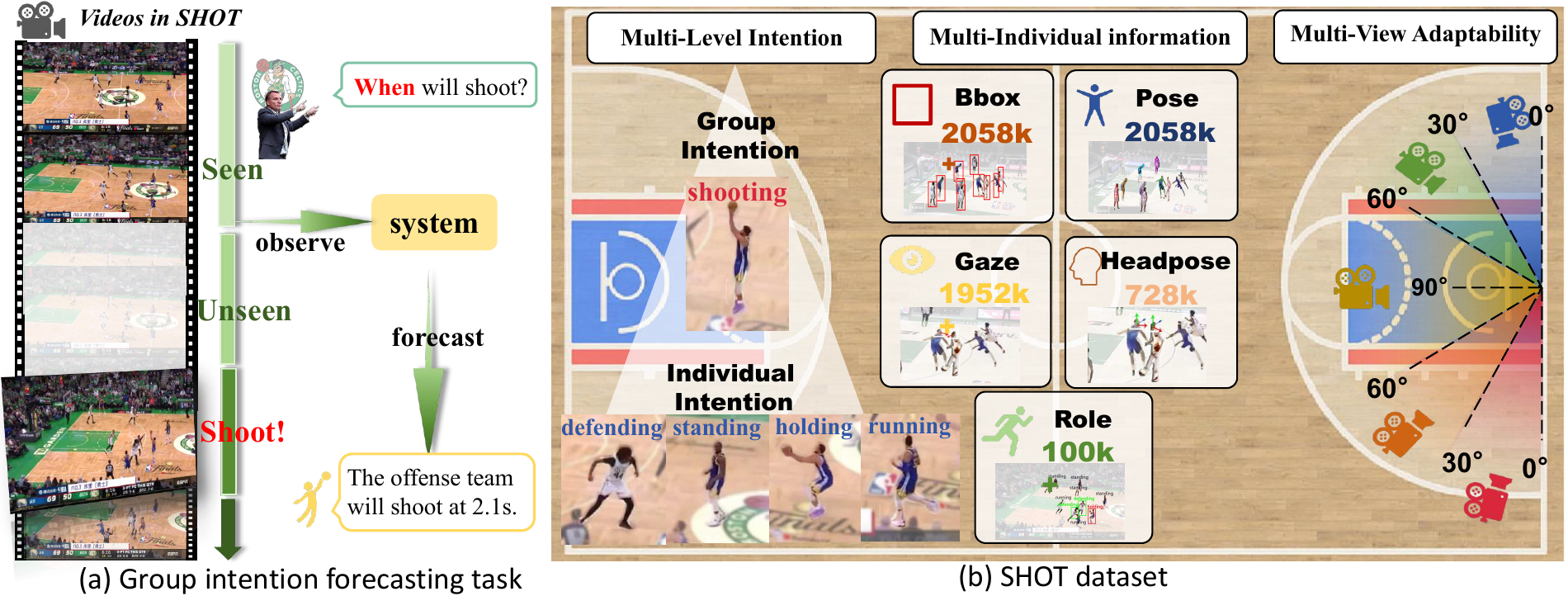}
  \caption{\textbf{Overview of the Group Intention Forecasting (GIF) task and the SHOT dataset.} 
	(a) Group Intention Forecasting task forecasts the occurrence time of group intentions by observing individual actions and interactions in early time. 
	(b) The SHOT dataset provides 5 camera views videos and is annotated with 6 multi-individual attributes to describe the multi-level intention.}
  \Description{}
  \label{fig:dataset}
\end{teaserfigure}

\maketitle
\section{Introduction}

Intention recognition has gained significant attention and has been applied in various domains, including human-computer interaction~\cite{WangWHYLZG023,PearceAB23} and intelligent security systems~\cite{CordobaJABSSPK23}.
However, existing studies~\cite{XuMZC22,0014LH0HZ22,HamKJM23,ZhouXLZZWG24,ZhouWZWYZGC24,10.1145/3664647.3681113,hu2025span} mainly focus on \emph{individual intention}, \textit{i.e.,} the goals of a single person, while largely overlooking the collective intentions of groups.
This narrow scope limits real-world applicability, where understanding group-level dynamics and multi-agent interactions is essential.
For example, in coordinated crimes, a single individual may appear to act innocuously, yet plays a critical role in a collective plan. Without modeling group intention, such actions may seem benign or go unnoticed, hindering effective crime prevention.
Such intention behind coordinated crimes often emerges through a complex yet observable process before it fully materializes. Early perception of such intentions is critical for effective prevention.
However, current intention recognition tasks allow little reaction time before intentions become explicit, reducing the chance for timely intervention in high-stakes situations.

To address these limitations, we introduce \textbf{\emph{group intention}}, which represents the shared goals manifested through coordinated behaviors among individuals.
This concept operates at a higher level than \emph{individual intention}.
For example, in team sports, each player's actions (running, defending, etc.) reflect individual intentions, but their coordination toward scoring embodies the team’s \emph{group intention}.
Building on this concept, we propose the \textbf{\emph{Group Intention Forecasting}} (GIF) task, which aims to forecast the timing of group intentions by observing early-stage individual actions and interactions.
In basketball, for example, defense teams must anticipate when the offense will shoot in order to disrupt their play within the pre-shot window.
By forecasting group intentions early, GIF enables timely decisions beyond what individual-level analysis can reveal, benefiting domains like sports strategy, public safety, and intelligent systems.

However, existing datasets do not support GIF.
Datasets for intention recognition~\cite{ijcai/WeiXZZ17,cvpr/WeiLSZZ18,ZhangXWZZT22} focus on individuals and lack group-level annotations, while group activity datasets~\cite{yan2020social,cvpr/IbrahimMDVM16,cvpr/RamanathanHAG0F16} emphasize explicit group actions rather than emerging intentions.
To fill this gap, we introduce the \textbf{\emph{SHOT dataset}}, a large-scale dataset tailored for GIF. It consists of 1,979 basketball clips from 5 views and is annotated with 6 types of individual attributes.
SHOT has three essential features for GIF:
(1) Multi-Individual Information: Captures fine-grained cues (\textit{i.e.,}  bbox, pose, gaze, head orientation, role, velocity) for early-stage analysis.
(2) Multi-View Adaptability: Multiple views reduce occlusions and enable precise tracking of interactions.
(3) Multi-Level Intention: Captures both individual behaviors and their coordination via role annotations.

In addition, existing algorithms lack mechanisms to forecast the timing of group intentions.
Traditional group activity methods focus on ``what is happening'', while action prediction methods target individuals and overlook group interactions.
To address this challenge, we propose the \textbf{\emph{GIFT}} (\textbf{G}roup \textbf{I}ntention \textbf{F}orecas\textbf{T}er) \textbf{\emph{framework}}. GIFT extracts heterogeneous player features from seen frames and uses an encoder-decoder architecture with Spatio-Temporal Graph Convolutional Networks to model interaction dynamics. It forecasts future player features and identifies the moment a group intention, such as a basketball shot, occurs.

Our contributions are threefold:
\begin{itemize}
\item \textbf{A Novel Task:} We define Group Intention Forecasting (GIF) to capture shared objectives emerging from coordinated group behavior, enabling early-stage analysis and intervention.
\item \textbf{A Comprehensive Dataset:} We release SHOT, the first dataset for GIF, featuring 1,979 clips and 6.8M annotations, with multi-individual information, multi-view adaptability, and multi-level intention.
\item \textbf{An Effective Baseline:} We introduce \textsc{GIFT} as a foundation baseline. Experiments demonstrate the value of SHOT and the effectiveness of our method.
\end{itemize}

\section{Related Work}
\subsection{Related Datasets} 

We compare two types of datasets: individual intention datasets and sports analysis datasets (see Table~\ref{table1}).
Individual intention datasets~\cite{ijcai/WeiXZZ17,cvpr/WeiLSZZ18,ZhangXWZZT22,10.1145/3664647.3681113} typically focus on short-term, individual intention, often derived from a single observation view and relatively simple scenarios, where intentions can be inferred from successive actions. 
Sports analysis datasets~\cite{hsieh2019,mm/ChenLWCTL15,tcsv/KongPHHW22} center on immediate strategies~\cite{tcsv/KongPHHW22} and trajectories~\cite{hsieh2019}, neglecting player characteristics and group interactions.
~\cite{yan2020social,cvpr/IbrahimMDVM16,cvpr/RamanathanHAG0F16} focus on overall group activities rather than the progressively emerging intentions of the group.
Most of the above datasets rely on broad, traditional annotations without task-specific details.
In contrast, our SHOT dataset offers 5 camera views of information and links dynamic individual roles with shared goals, supporting fine-grained analysis of group behavior and providing necessary cues for forecasting group intentions.

\subsection{Intention Recognition} 
Recent work~\cite{WeiXZZ17,WeiLSZZ18,FangWRWSLFH20,LiuACLSGN20} has successfully applied intention recognition to individual human intent by predicting future actions from past sequences.
Other studies~\cite{XuHXWLL21, XuHWLLZ22} leverage nonverbal cues such as gaze to improve recognition accuracy.
ICCA~\cite{ZhouWZWYZGC24} introduces temporal modeling and evolutionary learning to detect the timing and pattern of intentions.
Multi-modal approaches~\cite{WanyanYMX23,ZhouXLZZWG24} are also gaining traction in pedestrian intention prediction~\cite{ZhangTD23}.
However, these methods primarily address individual intention, neglecting group interactions and thus limiting their ability to capture the complexity of group activity.
Currently, no method specifically targets group intention forecasting. To fill this gap, our GIFT baseline leverages spatio-temporal features to integrate both individual and group-level information with the SHOT dataset’s rich annotations.

\section{Key Concept and Task}
\subsection{Group Intention}
Group intention is defined as the shared goals manifested through coordinated actions among individuals.

\paragraph{Group Intention vs. Individual Intention}
From the perspective of the target subject, group intention refers to the collective goal of a group, whereas individual intention pertains to the personal goal of a single person.
From the perspective of the formation process, group intention emerges from individual actions and bidirectional interactions among group members, while individual intention arises from individual actions and the unidirectional influence of the environment on the individual.
From the perspective of occurrence characteristics, group intention occurs as a single event at a given time, whereas individual intentions within a group may occur simultaneously and even identically across different individuals.

\paragraph{Group Intention vs. Group Activity}
From the perspective of the formation process, group intention emphasizes specific individual actions and interactions, whereas group activity concerns a more abstract representation of group behavior.
From the perspective of the manifestation period, group intention emerges progressively over time, while group activity is directly observable.

\subsection{Group Intention Forecasting Task}
\subsubsection{Task Definition}
Group Intention Forecasting (GIF) aims to forecast the timing of group intentions by observing early frames that capture diverse individual actions and interactions among multiple individuals before the intention becomes observable, which is shown in Figure~\ref{fig:dataset} (a).
Given a video clip segmented into $T$ frames, with an observed frame length \(\tau\) represented as \(\{f_1, f_2, ..., f_\tau, ..., f_T\}\), where \(\{f_1, ..., f_\tau\}\) denotes the early seen frames and \(f_\tau\) represents the occurrence frame of the group intention within the clip, the goal is to forecast \(f_\tau\) based on the early seen frames.

\subsubsection{Task Significance}

The significance of the GIF task is twofold.
\paragraph{Group-Level Analysis}
GIF is the first task dedicated to recognizing shared goals within a group and leveraging the unique insights they provide.
Group intention reveals critical patterns that individual intention cannot.
For instance, in a coordinated criminal act, a single perpetrator may carry out a minor task. If only individual intentions are considered, such actions may seem benign or go unnoticed, hindering effective crime prevention.

\paragraph{Early-Stage Forecasting}
GIF emphasizes early forecasting by accounting for the marginal effect of decision-making costs.
In group contexts, the relationship between time and resource expenditure is nonlinear; early intervention typically demands exponentially fewer resources.
For example, early in a match, repositioning a single player may be sufficient to disrupt the opponent’s tactical chain. By mid-game, repositioning may involve multiple players and require significant adjustments. In late-game situations, a timeout may be necessary to reorganize the strategy.
Furthermore, the earlier an intention is forecasted, the higher the likelihood of achieving an optimal response.

\section{The SHOT Dataset}

\begin{figure}
	\centering
	\includegraphics[width = \linewidth]{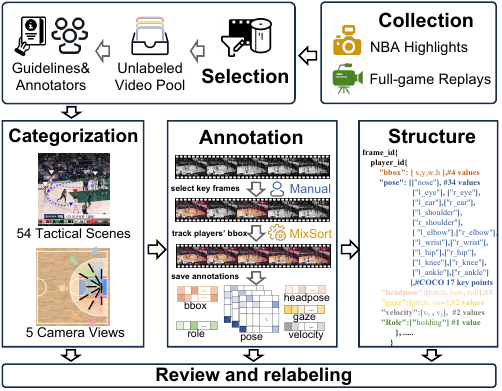}
	\caption{\textbf{Dataset pipeline overview.} \textbf{Collection:} videos are sourced from NBA highlights and full-game replays, then compiled into an unlabeled pool. \textbf{Categorization:} clips are classified by camera view and tactical type. \textbf{Annotation:} features are labeled manually or via tracking models. \textbf{Structure:} video annotations are stored in a JSON file with this structure. \textbf{Review:} annotations are reviewed and relabeled as needed.}
	\label{fig:pipeline}
\end{figure}

\subsection{Dataset Construction} 
The construction of SHOT follows a well-organized pipeline spanning from initial video selection to the final integration of enriched annotations. We employ 30 annotators provided with standardized guidelines to ensure consistency and quality. The annotation process involves annotators working 3 hours daily for 2 weeks. To maintain high standards, we implement several mechanisms, including training annotators with domain-specific knowledge, conducting rigorous pre-tests, preparing reference materials, and cross-validating annotations. The entire data preparation and annotation process is illustrated in Figure~\ref{fig:pipeline} and is described as follows.

\subsubsection{Data Collection} SHOT is sourced from Tencent Sports. We analyze NBA basketball games from the past five years, obtaining high-resolution (1080p) video footage for further processing. Figure~\ref{fig:dataset}(b) presents screenshots of an annotated sample. 

\subsubsection{Data Selection} This step involves carefully reviewing the video footage to identify and extract shooting clips, ensuring that each clip includes all players on the court. Each segment is then cropped and saved for subsequent analysis.

\subsubsection{Categorization} 
\paragraph{Views} 
Video clips are classified into 5 views based on camera angles spaced $30^{\circ}$ apart, with the central view covering $60^{\circ}$. 
\paragraph{Tactical Scenes} 
Videos are divided into 54 categories based on tactics. 
Each clip is categorized based on 4 key dimensions that reflect the complexity of basketball strategies. \textit{Passing Frequency} divides scenes into No-Pass (no pass made before the shot), One-Pass (one pass before the shot), and Multi-Pass (multiple passes before the shot). \textit{P\&R Frequency} distinguishes between No-P\&R (no pick-and-roll), One-P\&R (one pick-and-roll), and Multi-P\&R (multiple pick-and-rolls). \textit{Drive} identifies whether a drive (aggressive movement toward the basket) occurs, or if it is absent. \textit{Shooting Method} categorizes the final shot into Shoot, Layup, and Dunk.

\subsubsection{Annotations} 
\paragraph{Keyframe Selection} Consecutive frames are extracted from video clips with average fps of 25, and keyframes marking transitions between temporal phases are selected including the first frame of the video, the first frame when the basketball player receives the ball, the first frame when the basketball player prepares to shoot, the first frame when the basketball player begins the shot, and the first frame when the basketball player's shooting motion ends.

\paragraph{Individual Features Annotation} For each keyframe, we assign player IDs sequentially from bottom to top across the court, with offensive players labeled 1--5 and defensive players labeled 6--10. For multi-view data, we ensure consistent ID alignment across views. We manually annotate the bounding boxes (BBoxes) of all players in each keyframe. Additionally, we classify each player's role into one of seven categories: standing, running, defending, holding, shooting, laying-up, or dunking. For intermediate frames, we use MixSort~\cite{CuiZZYWW23} with annotated BBoxes as reference points to track players across frames and obtain BBoxes for all frames. Once tracking is completed, we use RTMO~\cite{abs-2312-07526} to compute and save pose information for each player, integrating the previously obtained BBoxes into the model for more accurate pose estimation. We then use L2CS-Net~\cite{AbdelrahmanHKAD23} to capture and store gaze data, and DirectMHP~\cite{abs-2302-01110} to calculate and save headpose information for each player.

\subsubsection{Structure} Each video annotation is stored in a \textsc{JSON} file, containing frame\_id, player\_id, and the annotation. Features retain the dimensional definitions: 
\textbf{Bbox:} $x$, $y$, $h$, $w$.
\textbf{Pose:} 17 \textsc{COCO} keypoints, each with $x$ and $y$ (34 values).
\textbf{Gaze:} $pitch$, $yaw$.
\textbf{Headpose:} $pitch$, $yaw$, $roll$.
\textbf{Velocity:} $v_x$, $v_y$, the coordinate change between frames divided by time. 
The velocity of an object in the \(x\)- and \(y\)-directions can be computed by considering the change in the bounding box coordinates between consecutive frames divided by the time interval. Specifically, for each direction, the change in the \(x\)-coordinate (\( \Delta x \)) and the \(y\)-coordinate (\( \Delta y \)) of the object's bounding box between the current and previous frames is calculated, and then divided by the time elapsed between the frames. This can be mathematically expressed as:
\[
v_x = \frac{x_{\text{current}} - x_{\text{previous}}}{\Delta t}, \quad v_y = \frac{y_{\text{current}} - y_{\text{previous}}}{\Delta t}
\]
Where 
\( v_x \) and \( v_y \) are the velocities along the \(x\)- and \(y\)-axes,
\( x_{\text{current}} \) and \( y_{\text{current}} \) are the coordinates of the bounding box in the current frame,
\( x_{\text{previous}} \) and \( y_{\text{previous}} \) are the coordinates of the bounding box in the previous frame,
\( \Delta t \) is the time difference between the two frames.
For the first frame, the velocity is defined as 0.

\subsubsection{Review and Relabeling} To thoroughly ensure the quality of our dataset annotations, we carefully conduct a second round of review by experienced annotators, who cross-check the initial annotations and correct any inaccuracies.

\begin{table*}
	\centering
	\footnotesize
	\setlength{\tabcolsep}{2pt}
        \caption{\textbf{Comparison of the proposed SHOT dataset with existing datasets.} SA: Sports Analysis, II: Individual Intention, GIF: Group Intention Forecasting.}
        \scalebox{1.15}{
	\begin{tabular}{clcccccccccccccc} 
	\toprule[1.1pt]
	\multirow{2}[2]{*}{Task} & \multirow{2}[2]{*}{Dataset} & \multirow{2}[2]{*}{\begin{tabular}{c} Group \\ Intention \end{tabular}} & \multirow{2}[2]{*}{\begin{tabular}{c} Early time \\ Information \end{tabular}} & \multicolumn{2}{c}{Group Information} & \multicolumn{6}{c}{Individual Information} & \multirow{2}{*}{Video} & \multirow{2}{*}{Frame} & \multirow{2}{*}{Time} & \multirow{2}{*}{Annotation} \\
	\cmidrule(lr){5-6} \cmidrule(lr){7-12}
	& & & & Multi-View & Tactic & ID & BBox & Pose & Gaze & Headpose & Role \\ 
	\midrule
	\multirow{12}{*}{SA}& \textsc{CAD~\cite{5457461}} & - & - & $\times$ & $\times$ & $\times$ & - & $\times$ & $\times$ & $\times$ & $\times$ & 44 & - & - & - \\
	& \textsc{Volleyball~\cite{cvpr/IbrahimMDVM16}} & - & - & $\times$ & $\times$ & $\times$ & - & $\times$ & $\times$ & $\times$ & $\times$ & 4,830 & 1,525 & 1h34min & 15K \\
	& \textsc{NBA~\cite{yan2020social}} & - & - & $\times$ & $\times$ & $\times$ & - & $\times$ & $\times$ & $\times$ & $\times$ & 257 & 14K & 1h30min & 120K \\
	& \textsc{VolleyTactic~\cite{9727174}} & - & -& $\times$ & 6 & 6 & 36K & $\times$ & $\times$ & $\times$ & $\times$ & 4,960 & - & - & 36K \\ 
	 & \textsc{UCF SPORTS~\cite{soomro2015action}} & - & - & $\times$ & $\times$ & $\times$ & 9,580 & $\times$ & $\times$ & $\times$ & $\times$ & 150 & 9,580 & 16min & 9,580 \\
	& \textsc{FSN~\cite{Yu_2018_CVPR}} & - & - & $\times$ & $\times$ & $\times$ & - & $\times$ & $\times$ & $\times$ & $\times$ & 2,000 & - & 2h38min & - \\
	& \textsc{MULTISPORTS~\cite{li2021multisportsmultipersonvideodataset}} & - & -& $\times$ & $\times$ & 4 & 900K & $\times$ & $\times$ & $\times$ & $\times$ & 3,200 & 37K & - & 900K \\
	& \textsc{FINEDIVING~\cite{xu2022finediving}} & - & - & $\times$ & $\times$ & $\times$ & - & $\times$ & $\times$ & $\times$ & $\times$ & 3,000 & - & - & - \\
	& \textsc{NBA GAME~\cite{hauri2022groupactivityrecognitionbasketball}} & - & - & $\times$ & $\times$ & $\times$ & - & $\times$ & $\times$ & $\times$ & $\times$ & 181 & - & - & - \\
	& \textsc{SportsMOT~\cite{cui2023sportsmot}} & - & - & $\times$ & $\times$ & $\times$ & 160K & $\times$ & $\times$ & $\times$ & $\times$ & 240 & 150K & 1h40min & 160K \\
	& \textsc{SoccerNet~\cite{9857224}} & - & - & 4 & $\times$ & 5 & 360K & $\times$ & $\times$ & $\times$ & $\times$ & 201 & 220K & 1h41min & 360K \\
	& \textsc{FineSports~\cite{10657871}} & - & - & $\times$ & $\times$ & $\times$ & 123K & $\times$ & $\times$ & $\times$ & $\times$ & 10,000 & - & - & 139K \\
	& \textsc{SportsHHI~\cite{wu2024sportshhi}} & - & -& $\times$ & $\times$ & 10 & 118K & $\times$ & $\times$ & $\times$ & $\times$ & 160 & 96.5K & 1h4min & 169K \\
	\midrule
	\multirow{3}{*}{II} & \textsc{3D Attention~\cite{ijcai2017p180}} & $\times$ & $\times$ & $\times$ & -  & $\times$ & - & $\times$ & $\times$ & $\times$ & $\times$ & 150 & - & - & - \\
	& \textsc{TIA~\cite{8578809}} & $\times$ & $\times$ & $\times$ & -  & $\times$ & - & $\times$ & $\times$ & $\times$ & $\times$ & 809 & 330K & - & 330K \\
	& \textsc{MIntRec~\cite{10.1145/3503161.3547906}} & $\times$ & $\times$ & $\times$ & -  & $\times$ & - & $\times$ & $\times$ & $\times$ & $\times$ & 2,224 & - & 1h28min & 12K \\
	\midrule
    \rowcolor{gray!20}
	{\textbf{GIF}} & {\textbf{SHOT}} & {\textbf{\checkmark}} & {\textbf{\checkmark}} & {\textbf{5}} & {\textbf{54}} & {\textbf{10}} & {\textbf{2.1M}} & {\textbf{\checkmark}} & {\textbf{\checkmark}} & {\textbf{\checkmark}} & {\textbf{7}} & {1,979} & {214K} &{2h16min} & {\textbf{6.9M}} \\
	\bottomrule[1.1pt]
	\end{tabular}
	}
	\label{table1}  
\end{table*}

\subsection{Dataset Statistics}

SHOT has a larger data scale including up to 1,979 video clips captured from 5 different camera views, totaling 2 hours and 16 minutes of footage. The dataset features 6 types of individual annotations, encompassing approximately 2.1 million player positions, pose instances and velocity data, 2.0 million gaze annotations, 730,000 headpose annotations, and over 100,000 role labels for the ten players on the court. 
Table~\ref{table1} presents the statistics for SHOT, comparing it with other related video datasets.
Compared to sports analysis datasets, our dataset stands out not only due to its larger scale, but also because we have annotated finer-grained details like player poses, gazes and headpose which enhance the dataset's utility for understanding sports games. Furthermore, in comparison to intention-related datasets, our dataset is unique in that it explicitly annotates group intention and we emphasize early stages of manifestation of intention.

\subsection{Dataset Characteristics}

Our dataset addresses the complex and challenging task of forecasting group intention.

Unlike conventional datasets that focus primarily on individual intentions or simple group activities, SHOT involves analyzing individual actions and understanding how individuals within a group align to achieve a shared goal, which is more complicated in the early time analysis due to the dynamic nature of team sports like basketball.

\paragraph{Multi-Individual Information}
In early-stage analysis, the scarcity of explicit key signs presents a significant challenge, making it essential to capture rich, multi-dimensional information at the individual level. Since group intentions are not yet fully revealed in the early stages of an interaction, relying solely on coarse or high-level observations is often insufficient. Therefore, SHOT addresses the challenge by providing detailed annotations of individual attributes, such as bbox, pose, gaze directions, head orientations and roles, which serve as critical indicators of emerging intentions. 
These fine-grained features allow for a more comprehensive understanding of how each player's behavior contributes to the subtle formation of group intention.

\paragraph{Multi-View Adaptability} 
Early-stage analysis often suffers from occlusion issues in single-view datasets, where critical player actions may be hidden by other players or objects on the court. The multi-view method mitigates this problem by providing alternative viewpoints, ensuring that occluded players or actions are still observable from different angles. This results in a more robust and accurate analysis of player intentions, even when certain movements or interactions are not visible from a single perspective~\cite{zhong2025,liu2023dual}.

\paragraph{Multi-Level Intention}

The group shooting intention emerges from the interaction of individual behaviors, involving both the group subject and the individual subject.
From the individual subject perspective, each player engages in distinct actions such as holding, running, or defending, reflecting their immediate decision-making and positioning relative to the ball and teammates.
From the group subject perspective, players' actions and interactions collectively contribute to the shooting intention.
The multi-level intention is represented through role annotation, encompassing both statuses reflecting the individual subject, such as running, and those reflecting the group subject, such as shooting.

\begin{figure*}
\centering
\includegraphics[width = \textwidth]{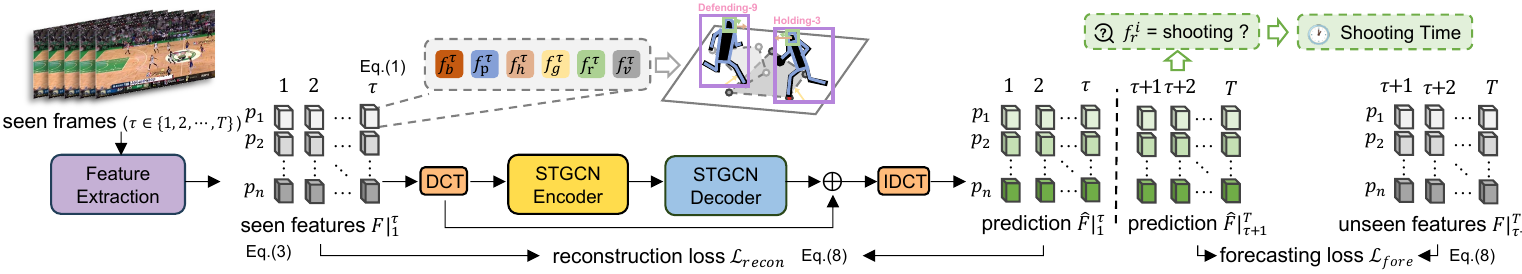} 
\caption{
Architecture of GIFT. GIFT extracts bounding box, pose, gaze, headpose, velocity, and role features from the $\mathbf{\tau}$ seen frames ($\mathbf{\tau \in {1, 2, \dots, T}}$). The STGCN Encoder models spatial and temporal patterns. The STGCN Decoder forecasts future features, from which the shooting role is identified to determine the frame number.}
\label{fig:baseline}
\end{figure*}

\section{The GIFT Method}
\subsection{Overview}
Given the first $\tau$ frames ($\tau \in \{1, 2, \dots, T\}$, manually set) of a video clip $X \in \mathbb{R}^{T \times H \times W \times C}$ as input ($T, H, W, C$ denote the length, height, width, and channels), the model forecasts the frame number involving shooting in the unseen segment.
The seen frames provide 6 individual features from our SHOT dataset: bounding box $f_b$, pose $f_p$, gaze $f_g$, headpose $f_h$, role $f_r$, and velocity $f_v$.

We propose GIFT (Group Intention ForecasTer) to solve this task, as shown in Figure~\ref{fig:baseline}.
By analyzing the $\tau$ seen frames, GIFT forecasts each player's features in the unseen frames and outputs the shooting time.

\subsection{Architecture}
GIFT employs an encoder-decoder architecture~\cite{xu2024learning}, comprising an STGCN~\cite{yu2017spatio} Encoder and an STGCN Decoder.

\subsubsection{Feature Extraction}
The six heterogeneous features are vectorized and concatenated into a single input vector per player.
For the $i$-th player ($i \in \{1, 2, \dots, n\}$): bounding box $f_b^i \in \mathbb{R}^4$, pose $f_p^i \in \mathbb{R}^{34}$, gaze $f_g^i \in \mathbb{R}^2$, headpose $f_h^i \in \mathbb{R}^3$, velocity $f_v^i \in \mathbb{R}^2$, and role $f_r^i \in \mathbb{R}^1$.
Roles are binarized: if role $\in$ \{standing, running, defending, holding\}, then $f_r^i = 0$; if role $\in$ \{shooting, laying-up, dunking\}, then $f_r^i = 1$.

The concatenated player feature $p_i$ is defined as:
\begin{equation} p_i = [f_b^i, f_p^i, f_h^i, f_g^i, f_v^i, f_r^i] \in \mathbb{R}^{46},\ i \in \{1, 2, \dots, n\} \end{equation}

The $k$-th frame feature $F_k$ is the concatenation of all $n$ player features:
\begin{equation} F_k = [p_1, p_2, ..., p_n],\ k \in \{1, 2, ..., T\} \end{equation}

Thus, the features from the $\tau$ seen frames are:
\begin{equation} F|_1^\tau = [F_1, F_2, ..., F_\tau],\ \tau \in \{1, 2, \dots, T\} \end{equation}

Before input to the network, $F|_1^\tau$ is embedded into a higher-dimensional space using a DCT~\cite{Mao_2021_ICCV,9009559} layer to obtain more informative representations.

\begin{table*}
	\centering
	\footnotesize
    
    \caption{Quantitative comparison of leading methods on SHOT. Best performances are highlighted in bold.}
    \scalebox{1.2}{
	\begin{tabular}{cccccccc}
	\toprule[1.1pt]
	\multirow{2}{*}{Method} & 
	\multirow{2}{*}{Venue} & 
	\multirow{2}{*}{Task} & 
	\multirow{2}{*}{Seen Frames} & 
	\multicolumn{3}{c}{\textit{Occurrence Evaluation}} & 
	\textit{Timing Deviation} \\
	& & & & Recall$\uparrow$ & Precision$\uparrow$ & F1-score$\uparrow$ & MAE$\downarrow$ \\
	\midrule
	\textsc{Actionformer}~\cite{zhang2022actionformer} & ECCV’22  & Localization & all & - & - & - & 26.7 \\
	\textsc{Tridet}~\cite{shi2023tridet} & CVPR’23  & Localization & all & - & - & - & 30.2 \\
	\textsc{Dydated}~\cite{10.1007/978-3-031-72952-2_18} & ECCV’24   & Localization & all & - & - & - & 28.9 \\
	\cmidrule{1-8}
	\rowcolor{gray!20}
    \textsc{GIFT} (Ours) & -  & Forecasting & 10 & \textbf{0.0080} & \textbf{0.1765} & \textbf{0.0153} & \textbf{15.6} \\
	\bottomrule[1.1pt]
	\end{tabular}
	}
	\label{exp1}
\end{table*}

\subsubsection{STGCN Encoder}
The STGCN Encoder captures spatial and temporal dynamics from the seen frames through four stacked blocks. Each block includes a Graph Convolutional Network (GCN)~\cite{kipf2016semi}, a temporal convolution, and a residual connection.
Spatial interactions $F_\text{spatial}$ are extracted by:
\begin{equation}
    F_{\text {spatial}}=GCN(F)
\end{equation}
where $F$ is the input feature vector. Temporal dependencies are captured by 2D convolution, followed by a residual connection:
\begin{equation}
    F_{\text{temporal}}=Conv2D(F_{\text{spatial}})
\end{equation}
\begin{equation}
    F_\text{encoder}=ReLU(F_{\text{temporal}}+Res(F_{\text{temporal}}))
\end{equation}

\subsubsection{STGCN Decoder}
The STGCN Decoder forecasts future features using encoder outputs and prior decoder outputs. It transforms latent representations from past to future, guided by conditional inputs and frequency-domain cues. The decoder mirrors the encoder structure and uses an inverse discrete cosine transform (IDCT) to generate temporally coherent predictions.

\subsection{Training Objective}
The loss $\mathcal{L}$ consists of reconstruction loss $\mathcal{L}_\text{recon}$ and forecasting loss $\mathcal{L}_\text{fore}$:
\begin{equation}
\mathcal{L} =\lambda_\text{recon}\mathcal{L}_\text{recon}+\lambda_\text{fore}\mathcal{L}_\text{fore}+\lambda_\text{const}\mathcal{L}_\text{const}\end{equation}
where $\mathcal{L}_\text{recon}$ trains the model to reconstruct seen frames, while $\mathcal{L}_\text{fore}$ targets future prediction and $\mathcal{L}_\text{const}$ is defined as in~\cite{xu2024learning}.
$\lambda_\text{recon}$, $\lambda_\text{fore}$ and $\lambda_\text{const}$ control their balance.
$\mathcal{L}_\text{recon}$ and $\mathcal{L}_\text{fore}$ include six feature components per player:
\begin{equation}
\begin{aligned}
	\mathcal{L}_i &= \lambda_1 \mathcal{L}_\mathrm{mse} \left(f_b^i, \hat{f}_b^i \right) + \lambda_2 \mathcal{L}_\mathrm{mse} \left(f_p^i, \hat{f}_p^i \right) + \lambda_3 \mathcal{L}_\mathrm{mse} \left(f_h^i, \hat{f}_h^i \right) \\
	&+ \lambda_4 \mathcal{L}_\mathrm{mse} \left(f_g^i, \hat{f}_g^i \right) + \lambda_5 \mathcal{L}_\mathrm{mse} \left(f_v^i, \hat{f}_v^i \right) + \lambda_6 \mathcal{L}_\mathrm{mse} \left(f_r^i, \hat{f}_r^i \right)
\end{aligned}
\label{eq6}
\end{equation}
where $\mathcal{L}_\mathrm{mse}$ denotes the mean squared error.
$\lambda_1, \lambda_2, \dots, \lambda_6$ balance each term's contribution.

\section{Experiments}
We conduct extensive experiments to investigate the following research question: Is our method effective in solving the Group Intention Forecasting task and achieving good performance compared to other methods?

\subsection{Implementation Details}
\paragraph{Evaluation Metrics}
 
The results of our Basketball Shooting Forecasting task involve frame labels, and we employ four evaluation metrics: Recall, Precision, F1-score and the mean absolute error (MAE). Precision refers to the proportion of all samples predicted to be in the positive class that are truly in the positive class; Recall refers to the proportion of all samples that are truly in the positive class that are correctly identified. The F1-Score centers on finding a balance between Precision and Recall. MAE measures the average deviation between predicted and ground truth shooting frames, providing valuable insights into temporal prediction accuracy.

\paragraph{Experimental Setup}

We randomly select $20\%$ of SHOT dataset as the test set, with the remaining $80\%$ used for training. During training, the dataset is further split into training and validation sets in a 4:1 ratio. In \textsc{GIFT}, we set the $\tau$ to 10. 

The initial base learning rate is $10^{-3}$, and weight decay is applied with a coefficient of $10^{-4}$. The feature embedding dimension is set to $128$, dropout rate is $0.1$, and the number of flow layers is $4$. The hyper-parameters $\{\lambda_{\text{recon}}, \lambda_{\text{fore}}, \lambda_{\text{const}}, \lambda_1, \lambda_2, \lambda_3, \lambda_4, \lambda_5, \lambda_6\}$ are set as $\{2, 0.01, 10, 0.1, 0.05, 0.001, 0.1, 10, 0.1\}$. We train for 100 epochs on a single A100 GPU.

\subsection{Main Results }

Table~\ref{exp1} compares \textsc{GIFT} with state-of-the-art (SOTA) methods using the MAE metric. Compared to Temporal Action Localization (TAL) models such as \textsc{TE-TAD}~\cite{10654957}, our method achieves a substantial improvement of 15.6 frames in MAE, indicating more precise shooting time predictions. While TAL models are effective at pinpointing isolated actions, they often fail to capture the complex group dynamics inherent in team sports like basketball. In contrast, \textsc{GIFT} incorporates individual player roles and inter-player coordination, resulting in more accurate modeling of collective behavior and improved temporal precision.

In forecasting shooting events, our method shows lower F1-scores due to the increased difficulty of the GIF task, which requires early prediction based on only the initial $\tau$ frames. 
However, in group contexts, earlier forecasting is more valuable, as intervention costs grow nonlinearly with time. 
Our approach thus emphasizes early-stage intention forecasting, aiming to support timely and efficient downstream decisions.

\section{Conclusion}
In this paper, we propose a novel task, Group Intention Forecasting (GIF). To address the data scarcity in this domain, we construct the large-scale SHOT dataset with multi-individual information, multi-view adaptability, and multi-level intentions.
We propose GIFT, a spatio-temporal encoder-decoder that models interaction dynamics from heterogeneous features to forecast future actions and group intention timing.
Our SHOT dataset focuses on addressing GIF task, but the rich annotations it contains also offer significant guidance for other tasks.
We hope the SHOT dataset will become a cornerstone for future research, not only in basketball shooting forecasting but also in broader tasks.

\bibliographystyle{ACM-Reference-Format}
\balance
\bibliography{shot-bib}

\clearpage

\end{document}